\documentclass{bmvc2k}
\usepackage{changepage}
\usepackage{tikz}
\usepackage{graphicx}
\usepackage{tikz}
\usepackage{pgfplots}
\usepgfplotslibrary{colormaps}
\pgfplotsset{compat=1.18}
\usepackage{booktabs}
\usepackage{makecell}
\usepackage{graphicx}
\usepackage{array}
\usepackage{tabularx}
\usepackage{amssymb}
\usepackage{multirow}
\usepackage{algorithm}
\usepackage{algpseudocode}
\title{MoEIoU: Rethinking Bounding-Box Regression as a Mixture of Experts}

\addauthor{Vinay Edula}{vinaykumarreddyedula@gmail.com}{1}
\addauthor{Priyanka Bagade}{pbagade@iitk.ac.in}{1}

\addinstitution{
 Department of Computer Science and Engineering\\
 Indian Institute of Technology Kanpur\\
 Kanpur, U.P, India
}

\runninghead{MoEIoU}{Rethinking Bounding-Box Regression as a Mixture of Experts}

\begin{document}

\maketitle

\begin{abstract}
Bounding-box regression is a fundamental component of object detection, playing a critical role in precise object localization. Existing Intersection-over-Union (IoU)-based loss functions extend the IoU objective by incorporating geometric penalties, such as center-distance and aspect-ratio mismatch, to improve bounding-box regression. However, these penalties typically remain fixed throughout training and do not account for the optimization dynamics in which predicted boxes initially exhibit large center-distance and shape errors, with later stages focusing on improving overlap with the ground truth. To address this limitation, we introduce MoEIoU, a mixture-of-experts based regression loss that jointly models overlap, center alignment, and aspect-ratio mismatch. MoEIoU aggregates these components using a log-sum-exp function, which emphasizes the dominant localization error while maintaining smooth contributions from other terms. Additionally, a curriculum-based weighting schedule is employed to prioritize correcting box position and shape in early training stages and improving overlap in later stages. We evaluated proposed MoEIoU on PASCAL VOC, HRIPCB, and MS COCO using multiple YOLO architectures, along with large-scale simulation experiments. It consistently outperforms standard and recent state-of-the-art losses, demonstrating faster convergence and improved localization accuracy. We further show that this adaptive aggregation improves existing IoU-based losses, yielding consistent gains and providing more effective optimization guidance for bounding-box regression in object detection frameworks. 

\end{abstract}

%-------------------------------------------------------------------------
%-------------------------------------------------------------------------

\begin{adjustwidth}{0.20in}{0.20in}

\section{Introduction}

Object detection has made substantial progress in recent years, with bounding-box regression remaining a critical component of modern detection systems~\cite{zhao2019objectdetectiondeeplearning}. Early detectors treated box prediction as a coordinate-wise regression problem, optimizing sum-squared or Smooth-$\ell_1$ losses over independent box parameters $(x,y,w,h)$~\cite{Redmon2016YOLO,Girshick2015FastRCNN,Huber1964}. Because these losses optimize coordinates independently, they do not directly capture box overlap with the ground truth, which is the main metric for detection performance~\cite{Lin2014COCO}. This motivated a shift toward IoU-based losses that treat the bounding box as a unified geometric object. UnitBox introduced $-\log(\mathrm{IoU})$, providing scale-invariant supervision that is directly aligned with detection evaluation metrics~\cite{Yu2016UnitBox}. 

\end{adjustwidth}

However, IoU yields zero gradients for non-overlapping boxes, stalling optimization in early training~\cite{Rezatofighi2019GIoU}. 

GIoU addressed this by incorporating the smallest enclosing box as an additional penalty\\~\cite{Rezatofighi2019GIoU}, but this term can encourage the predicted box to enlarge rather than move toward the ground truth, and it vanishes once the enclosing box equals the union of the two boxes~\cite{Zheng2020DIoU}. DIoU resolved this by introducing a normalized center-distance penalty that directly guides the predicted box toward the target even without overlap~\cite{Zheng2020DIoU}. CIoU further added an aspect-ratio mismatch term, jointly supervising overlap, position, and shape, and has since become the dominant regression loss in YOLO-based detectors~\cite{ZhengCIoU}. Later work focused on geometric refinements, for example, EIoU penalizes differences in box side lengths for more direct size supervision~\cite{Zhang2021FocalEIoU}; SIoU adds cost angle to guide the predicted box towards the target~\cite{Gevorgyan2022SIoU}; and Focal-EIoU incorporates a focal weighting scheme that increases the influence of high-quality anchors during training~\cite{Zhang2021FocalEIoU}.

\begin{figure}[t]
  \centering
  \includegraphics[
    width=0.78\textwidth,
    height=0.25\textheight,
    keepaspectratio
  ]{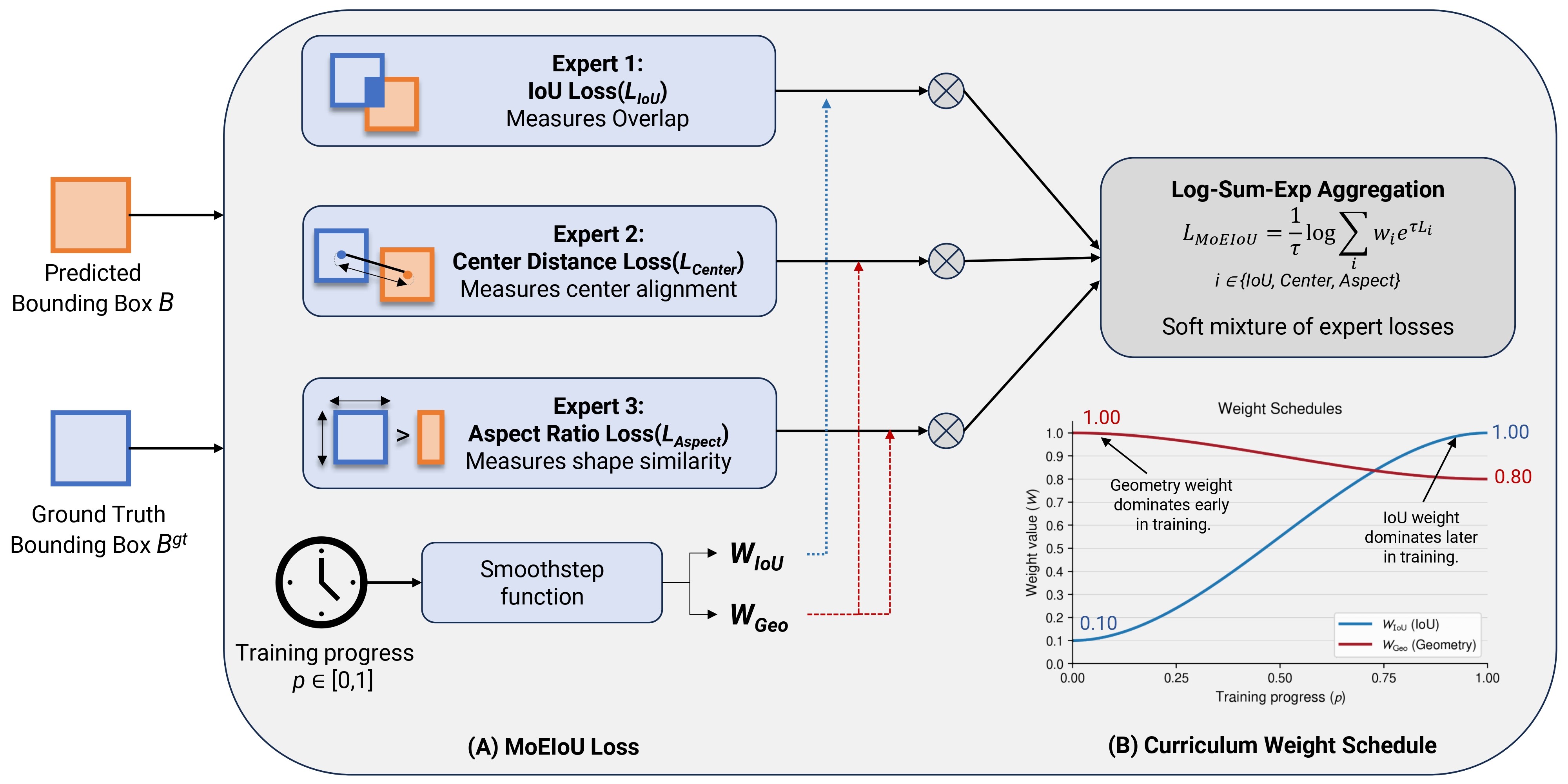}
  \caption{Overview of the proposed MoEIoU loss.}
  \label{fig:moeiou_overview}
\end{figure}

More recent losses focus on dynamically adjusting the contribution of individual predictions rather than refining fixed geometric terms ~\cite{Tong2023WIoU,Luo2024UIoU, Liu2024PIoU}. WIoU focuses regression on medium-quality boxes by suppressing unstable gradients from poor predictions and weak signals from already well-localized predictions~\cite{Tong2023WIoU}. PIoU and PIoU-v2 address the challenge that equal localization errors can have disproportionate effects on small versus large objects, introducing size-adaptive penalty factors and a non-monotonic attention mechanism to scale regression updates appropriately~\cite{Liu2024PIoU}. %UIoU applies a curriculum-style dynamic scaling strategy, early in training the loss emphasizes low-quality predictions to correct large errors, and as training progresses it shifts emphasis toward high-quality predictions to refine localization~\cite{Luo2024UIoU}.
UIoU applies a curriculum-style dynamic scaling strategy to refine localization by emphasizing on low-quality predictions to correct large errors, and as training progresses shifting toward high-quality predictions~\cite{Luo2024UIoU}.

Despite these advances, existing losses use \emph{fixed additive formulations}, keeping the contribution of each geometric term constant across training and predictions. In practice, however, different localization errors dominate at different stages. Early training benefits more from geometric cues such as center distance and aspect-ratio mismatch, while later stages require stronger overlap-based supervision for precise refinement. An adaptive regression loss that emphasizes the dominant localization error of each prediction is therefore a natural step toward more effective bounding-box optimization.
% Despite these advances, all existing losses combine their geometric components through \emph{fixed additive formulations}, keeping the relative contribution of each term constant throughout training and across predictions. In practice, however, predicted boxes are dominated by different localization errors at different stages. Early in training, predictions are far from the ground truth and benefit more from geometric signals such as center distance and aspect-ratio mismatch, which provide strong directional guidance. Later in training, predictions are already well-aligned and require stronger overlap-based supervision for precise boundary refinement. A regression loss that adapts to the dominant localization error of each prediction, rather than applying a fixed combination, is therefore a natural and important step toward more effective bounding-box optimization.

Utilizing this fact, we propose \textbf{MoEIoU}, a bounding-box regression loss that explicitly accounts for the dominant localization error in each prediction. The name reflects its mixture-of-experts~\cite{Jacobs1991AdaptiveMO} design, where each component —log-IoU, center distance, and aspect-ratio mismatch, which acts as an expert capturing a distinct aspect of localization quality. As illustrated in Figure~\ref{fig:moeiou_overview}, rather than adding these components with fixed weights, MoEIoU aggregates them using a \textbf{weighted log-sum-exp} function that behaves as a soft mixture-of-experts, naturally emphasizing the most informative signal for each prediction while retaining smooth contributions from the remaining terms. A \textbf{curriculum-based weighting strategy}~\cite{Curriculum} further modulates these contributions during training, placing greater emphasis on geometric terms in early stages when predictions are far from the target, and progressively increasing the influence of the overlap term as predictions improve and precise localization becomes the primary objective.

The main contributions of this work are:
\begin{itemize}
    \item We propose \textbf{MoEIoU}, a bounding-box regression loss that jointly models overlap, center alignment, and aspect-ratio mismatch.   
    \item We introduce a \textbf{weighted log-sum-exp(LSE) aggregation} that adaptively emphasizes the dominant localization error while preserving smooth contributions from other terms.
    \item We design a \textbf{curriculum weighting strategy} that shifts supervision from early geometric correction to later overlap refinement.
    \item We analyze MoEIoU through large-scale simulation experiments and compare its optimization behavior with existing IoU-based losses.
    \item We validate MoEIoU on PASCAL VOC, HRIPCB, and MS COCO using recent YOLO-based detectors, showing consistent localization gains.
\end{itemize}
%-------------------------------------------------------------------------

\section{Methodology}\label{sec:methodology}
This section presents the proposed MoEIoU loss for bounding box regression.
Specifically, Sec.~\ref{subsec:bbox_regression_components} defines the regression components, Sec.~\ref{subsec:curriculum_scheduling} introduces the training curriculum, Sec.~\ref{subsec:moeiou_formulation} presents the final MoEIoU objective, and Sec.~\ref{subsec:gradient_behavior_lse} analyzes its gradient behavior.

% The design of the loss is motivated by a simple observation: the relative importance of overlap and geometric alignment is not constant throughout training. In the early stages, predicted boxes are often far from their targets, making raw IoU less informative. At that stage, geometric signals such as center distance and shape mismatch provide more stable optimization signals. As training progresses and the predicted boxes become better aligned, overlap-based supervision becomes increasingly meaningful. The proposed loss is designed to reflect this progression explicitly.

\subsection{Bounding Box Geometry and Regression Components}
\label{subsec:bbox_regression_components}

Let the predicted bounding box be $b=(x_1,y_1,x_2,y_2)$ and the ground-truth box be
$b^{gt}=(x_1^{gt},y_1^{gt},x_2^{gt},y_2^{gt})$, where the two coordinate pairs denote the
top-left and bottom-right corners. The proposed MoEIoU models bounding-box regression through three
complementary components, a) an overlap term, b) a center-distance term, and c) an aspect-ratio
mismatch term, each capturing a distinct aspect of localization quality.

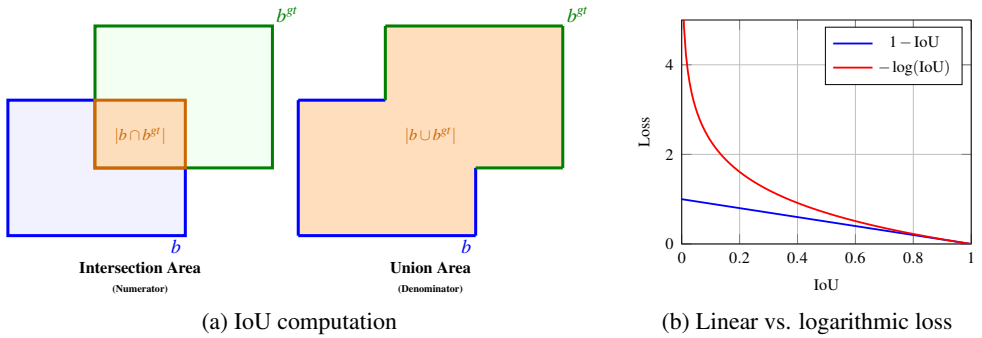
\begin{figure*}[t]
\centering

% ===================== LEFT: FULL IOU VISUALIZATION =====================
\begin{minipage}[t]{0.60\textwidth}
\centering
\resizebox{\linewidth}{!}{%
\begin{tikzpicture}[>=stealth, thick, scale=0.45, transform shape, every node/.style={font=\LARGE
}]

    % Common style for intersection/union labels
    \tikzset{
        ioumainlabel/.style={text=orange!80!black, font=\bfseries\Large}
    }

    % ---------------------------------------------------------
    % PANEL 1: INTERSECTION
    % ---------------------------------------------------------
    \draw[blue, fill=blue!5, very thick] (0, 0) rectangle (5.5, 4.2);
    \draw[green!50!black, fill=green!5, very thick] (2.7, 2.1) rectangle (8.2, 6.5);

    \fill[orange!30, opacity=0.85] (2.7, 2.1) rectangle (5.5, 4.2);
    \draw[orange!80!black, very thick] (2.7, 2.1) rectangle (5.5, 4.2);

    % Labels
    \node[text=blue, anchor=north east] at (5.5, 0) {$b$};
    \node[text=green!50!black, anchor=south west] at (8.2, 6.5) {$b^{gt}$};
    \node[ioumainlabel] at (4.1, 3.1) {$|b \cap b^{gt}|$};

    \node[align=center, font=\bfseries\Large] at (4.1, -1.3)
    {Intersection Area\\[-1pt] \small (Numerator)};

    % ---------------------------------------------------------
    % PANEL 2: UNION
    % ---------------------------------------------------------
    \begin{scope}[xshift=9cm]
        % Union fill
        \fill[orange!30, opacity=0.85]
            (0, 0) -- (5.5, 0) -- (5.5, 2.1) -- (8.2, 2.1) --
            (8.2, 6.5) -- (2.7, 6.5) -- (2.7, 4.2) -- (0, 4.2) -- cycle;

        % Visible outer border of box b (blue)
        \draw[blue, very thick] (0, 0) -- (5.5, 0);      % bottom
        \draw[blue, very thick] (0, 0) -- (0, 4.2);      % left
        \draw[blue, very thick] (0, 4.2) -- (2.7, 4.2);  % top-left visible part
        \draw[blue, very thick] (5.5, 0) -- (5.5, 2.1);  % right-bottom visible part

        % Visible outer border of box b^{gt} (green)
        \draw[green!50!black, very thick] (5.5, 2.1) -- (8.2, 2.1); % bottom-right visible part
        \draw[green!50!black, very thick] (8.2, 2.1) -- (8.2, 6.5); % right
        \draw[green!50!black, very thick] (2.7, 6.5) -- (8.2, 6.5); % top
        \draw[green!50!black, very thick] (2.7, 4.2) -- (2.7, 6.5); % left-top visible part

        % Labels
        \node[text=blue, anchor=north east] at (5.5, 0) {$b$};
        \node[text=green!50!black, anchor=south west] at (8.2, 6.5) {$b^{gt}$};
        \node[ioumainlabel] at (4.1, 3.1) {$|b \cup b^{gt}|$};

        \node[align=center, font=\bfseries\Large] at (4.1, -1.3)
        {Union Area\\[-1pt] \small (Denominator)};
    \end{scope}

\end{tikzpicture}%
}
\vspace{1mm}
\centerline{\small (a) IoU computation}
\end{minipage}
\hfill
% ===================== RIGHT: LOG-IOU GRAPH =====================
\begin{minipage}[t]{0.36\textwidth}
\centering
\resizebox{\linewidth}{!}{%
\begin{tikzpicture}
\begin{axis}[
    width=6cm,
    height=5cm,
    xlabel={IoU},
    ylabel={Loss},
    xmin=0, xmax=1,
    ymin=0, ymax=5,
    domain=0.001:1,
    samples=150,
    grid=both,
    legend style={font=\scriptsize, at={(0.97,0.97)}, anchor=north east},
    tick label style={font=\scriptsize},
    label style={font=\scriptsize}
]

\addplot[blue, thick, domain=0:1] {1 - x};
\addlegendentry{$1-\mathrm{IoU}$}

\addplot[red, thick, domain=0.001:1] {-ln(x)};
\addlegendentry{$-\log(\mathrm{IoU})$}

\end{axis}
\end{tikzpicture}%
}
\vspace{1mm}
\centerline{\small (b) Linear vs. logarithmic loss}
\end{minipage}

\caption{
IoU and logarithmic IoU penalty. 
(a) IoU is computed as the ratio between the intersection area and the union area of the predicted box $b$ and ground-truth box $b^{gt}$.
(b) Compared with the linear loss $1-\mathrm{IoU}$, the logarithmic loss $-\log(\mathrm{IoU})$ assigns a stronger penalty to low-overlap predictions.
}
\label{fig:iou_logiou_sidebyside}
\end{figure*}

\noindent\textbf{Overlap Term.}
The IoU metric measures the overlap between $b$ and $b^{gt}$ as
\begin{equation}
\mathrm{IoU}(b,b^{gt})=
\frac{|b\cap b^{gt}|}{|b\cup b^{gt}|},
\label{eq:iou_definition}
\end{equation}
%where $|b\cap b^{gt}|$ and $|b\cup b^{gt}|$ denote the intersection and union areas,respectively.
Although IoU is widely used for localization evaluation and regression,
the linear penalty $1-\mathrm{IoU}$ provides weak supervision for low-overlap predictions~\cite{Yu2016UnitBox}.
To strengthen the overlap penalty, we define the IoU term in logarithmic form:
\begin{equation}
T_{\mathrm{iou}}=-\log(\mathrm{IoU}+\varepsilon),
\label{eq:log_iou_term}
\end{equation}
where $\varepsilon>0$ avoids undefined values when IoU approaches zero. The negative
logarithm penalizes small IoU values more strongly, encouraging larger updates for poorly
aligned boxes while preserving the ordering induced by IoU. Figure~\ref{fig:iou_logiou_sidebyside}
illustrates IoU computation and the difference between linear and logarithmic overlap penalties.

\noindent\textbf{Center-Distance Term.}
Beyond overlap, bounding-box regression requires spatial alignment between the predicted and
ground-truth box centers. Let $(c_x,c_y)$ and $(c_x^{gt},c_y^{gt})$ denote the respective
centers. Their squared Euclidean distance is
\begin{equation}
\rho^2 = (c_x-c_x^{gt})^2 + (c_y-c_y^{gt})^2 .
\label{eq:center_distance}
\end{equation}
This displacement is normalized by the squared diagonal of the smallest enclosing box
covering $b$ and $b^{gt}$. Denoting the enclosing box width and height as $w_c$ and $h_c$,
the diagonal is $c^2 = w_c^2 + h_c^2$, and the center-distance term is defined as
\begin{equation}
T_{\mathrm{center}} = \sqrt{\frac{\rho^2}{c^2}+\varepsilon}.
\label{eq:center_term}
\end{equation}
This normalization makes the term scale-aware and ensures it remains informative even when
the two boxes have little or no overlap.

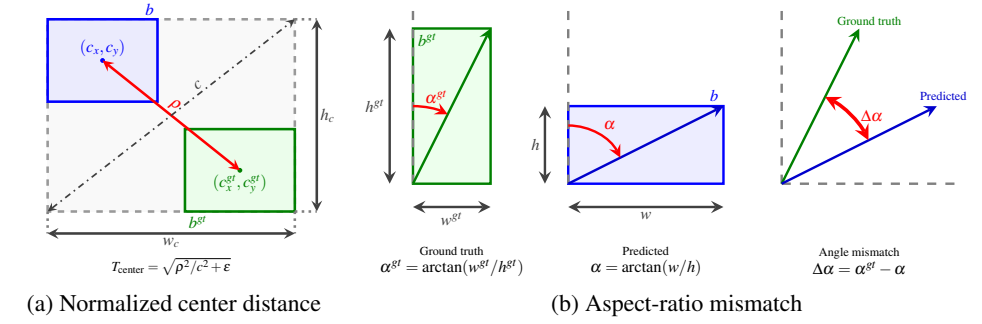
\begin{figure*}[t]
\centering

% Controls vertical alignment of the two panel captions
\newlength{\geomPanelHeight}
\setlength{\geomPanelHeight}{4.2cm}

% ===================== LEFT: CENTER DISTANCE =====================
\begin{minipage}[t]{0.35\textwidth}
\centering
\begin{minipage}[t][\geomPanelHeight][t]{\linewidth}
\centering
\resizebox{0.9\linewidth}{!}{%
\begin{tikzpicture}[>=stealth, very thick, scale=0.45, transform shape, every node/.style={font=\LARGE}]

    % Tight canvas: removed top empty padding
    \path[use as bounding box] (-1.0,-2.45) rectangle (10.2,7.15);

    % Coordinates
    \coordinate (P_BL) at (0, 4);
    \coordinate (P_TR) at (4, 7);
    \coordinate (P_C)  at (2, 5.5);

    \coordinate (GT_BL) at (5, 0);
    \coordinate (GT_TR) at (9, 3);
    \coordinate (GT_C)  at (7, 1.5);

    \coordinate (C_BL) at (0, 0);
    \coordinate (C_TR) at (9, 7);

    % Enclosing box
    \fill[gray!5] (C_BL) rectangle (C_TR);
    \draw[dashed, gray!80] (C_BL) rectangle (C_TR);

    % Predicted box
    \draw[blue, fill=blue!10, fill opacity=0.7] (P_BL) rectangle (P_TR);
    \node[text=blue, anchor=south east] at (P_TR) {$b$};
    \fill[blue] (P_C) circle (2.5pt)
    node[anchor=south] {$(c_x,c_y)$};

    % Ground-truth box
    \draw[green!50!black, fill=green!10, fill opacity=0.7] (GT_BL) rectangle (GT_TR);
    \node[text=green!50!black, anchor=north west] at (GT_BL) {$b^{gt}$};
    \fill[green!50!black] (GT_C) circle (2.5pt)
    node[anchor=north] {$(c_x^{gt},c_y^{gt})$};

    % Center distance
    \draw[<->, red, very thick] (P_C) -- (GT_C)
    node[pos=0.43, above right, sloped] {$\rho$};

    % Enclosing diagonal
    \draw[<->, darkgray, thick, dashdotted] (C_BL) -- (C_TR)
    node[pos=0.65, above left, sloped] {$c$};

    % Enclosing dimensions
    \draw[<->, darkgray] (0, -0.8) -- (9, -0.8)
    node[midway, below] {$w_c$};

    \draw[<->, darkgray] (9.8, 0) -- (9.8, 7)
    node[midway, right] {$h_c$};

    % Dotted connectors
    \draw[dotted, gray] (C_BL) -- (0, -0.8);
    \draw[dotted, gray] (9, 0) -- (9, -0.8);
    \draw[dotted, gray] (9, 0) -- (9.8, 0);
    \draw[dotted, gray] (C_TR) -- (9.8, 7);

    % Bottom label
    \node[align=center] at (4.5, -2.0)
    {\Large $T_{\mathrm{center}}=\sqrt{\rho^2/c^2+\varepsilon}$};

\end{tikzpicture}%
}

{\small (a) Normalized center distance}
\end{minipage}
\end{minipage}
\hfill
% ===================== RIGHT: ASPECT RATIO =====================
\begin{minipage}[t]{0.62\textwidth}
\centering
\begin{minipage}[t][\geomPanelHeight][t]{\linewidth}
\centering
\resizebox{\linewidth}{!}{%
\begin{tikzpicture}[>=stealth, thick, scale=0.45, transform shape, every node/.style={font=\large}]

    % Tight canvas: removed top empty padding
    \path[use as bounding box] (-1.0,-2.45) rectangle (14.6,4.65);

    % Ground-truth box
    \coordinate (GT_BL) at (0, 0);
    \coordinate (GT_TR) at (2, 4);

    \draw[green!50!black, fill=green!10, fill opacity=0.7] (GT_BL) rectangle (GT_TR);
    \node[text=green!50!black, anchor=north west] at (0, 4) {$b^{gt}$};

    \draw[dashed, gray] (GT_BL) -- (0, 4.5);
    \draw[->, thick, green!50!black] (GT_BL) -- (GT_TR);

    \draw[<->, darkgray] (0, -0.6) -- (2, -0.6)
    node[midway, below] {$w^{gt}$};

    \draw[<->, darkgray] (-0.6, 0) -- (-0.6, 4)
    node[midway, left] {$h^{gt}$};

    \draw[->, red, thick] (0, 2) arc (90:63.43:2)
    node[midway, right=4pt, above] {$\alpha^{gt}$};

    \node[align=center] at (1, -2.0)
    {\small Ground truth\\[-1pt]$\alpha^{gt}=\arctan(w^{gt}/h^{gt})$};

    % Predicted box
    \coordinate (P_BL) at (4, 0);
    \coordinate (P_TR) at (8, 2);

    \draw[blue, fill=blue!10, fill opacity=0.7] (P_BL) rectangle (P_TR);
    \node[text=blue, anchor=south east] at (P_TR) {$b$};

    \draw[dashed, gray] (P_BL) -- (4, 4.5);
    \draw[->, thick, blue!80!black] (P_BL) -- (P_TR);

    \draw[<->, darkgray] (4, -0.6) -- (8, -0.6)
    node[midway, below] {$w$};

    \draw[<->, darkgray] (3.4, 0) -- (3.4, 2)
    node[midway, left] {$h$};

    \draw[->, red, thick] (4, 1.5) arc (90:26.56:1.5)
    node[midway, above right] {$\alpha$};

    \node[align=center] at (6, -2.0)
    {\small Predicted\\[-1pt]$\alpha=\arctan(w/h)$};

    % Angle mismatch
    \coordinate (O) at (9.5, 0);

    \draw[dashed, gray] (O) -- ++(0, 4.5);
    \draw[dashed, gray] (O) -- ++(4.5, 0);

\draw[->, thick, green!50!black] (O) -- ++(63.43:4.47)
node[pos=0.97, above right=2pt,  xshift=-20pt, text=green!50!black] {\small Ground truth};

\draw[->, thick, blue!80!black] (O) -- ++(26.56:4.47)
node[pos=0.93, below right=2pt, yshift=20pt, xshift=-10pt, text=blue!80!black] {\small Predicted};

\draw[<->, red, very thick] (O) ++(26.56:2.5) arc (26.56:63.43:2.5)
    node[midway, right=4pt] {$\Delta\alpha$};

    \node[align=center] at (11.5, -2.0)
    {\small Angle mismatch\\[-1pt]$\Delta\alpha=\alpha^{gt}-\alpha$};

\end{tikzpicture}%
}

{\small (b) Aspect-ratio mismatch}
\end{minipage}
\end{minipage}

\caption{
Geometric components used in MoEIoU.
(a) The center-distance term measures the displacement $\rho$ between predicted and ground-truth box centers and normalizes it by the enclosing-box diagonal $c$.
(b) The aspect-ratio term~\cite{ZhengCIoU} represents each box shape by an angle and penalizes the angular mismatch between predicted and ground-truth aspect ratios.
}
\label{fig:geometric_components}

\end{figure*}

\noindent\textbf{Aspect-Ratio Term.}
To capture shape mismatch independently of absolute box scale, we use a CIoU based aspect-ratio term~\cite{ZhengCIoU}
based on the angular difference between the predicted and ground-truth width-height ratios:
\begin{equation}
T_{\mathrm{aspect}} =
\frac{4}{\pi^2}
\left(
\arctan\left(\frac{w^{gt}}{h^{gt}}\right)
-
\arctan\left(\frac{w}{h}\right)
\right)^2 .
\label{eq:aspect_term}
\end{equation}
Since this term depends only on $w/h$ and $w^{gt}/h^{gt}$, it penalizes shape discrepancy
without conflating it with position or size errors. Figure~\ref{fig:geometric_components}
illustrates both geometric components.

\subsection{Curriculum Scheduling}\label{subsec:curriculum_scheduling}

A key idea behind the proposed method is that different parts of the loss function are useful at different stages of training. During the early stages, predicted bounding boxes are typically coarse and may be significantly misaligned with the ground-truth boxes, sometimes even exhibiting no overlap. In such cases, relying on overlap-based supervision can lead to weak or unstable optimization signals. Geometric terms, such as center alignment and shape mismatch, remain informative even under large localization errors and therefore provide a more reliable guide for improving predictions. As training progresses and the predicted boxes become more closely aligned with the target objects, overlap-based supervision becomes increasingly meaningful. At that point, the optimization process should gradually place greater emphasis on improving the IoU between predicted and ground-truth boxes.

To model this transition smoothly, a scalar training-progress variable $p \in [0,1]$ is defined, where $p=0$ corresponds to the beginning of training and $p=1$ corresponds to the end. This progress value is then passed through a smoothstep scheduling function~\cite{smoothstep}:
\begin{equation}
s = p^2(3 - 2p).
\end{equation}

This function is monotonic, bounded in $[0,1]$, and differentiable, making it well suited for curriculum-style weighting~\cite{Curriculum}. It changes gradually near the endpoints and more actively in the middle, thereby avoiding abrupt shifts in the optimization objective. 
Using this schedule, the influence of the IoU component gradually increases during training,
\begin{equation}
w_{\mathrm{iou}} = 0.10 + 0.90s,
\label{eq:iou_weight}
\end{equation}
while the geometric components are kept active throughout training with a slight reduction in weight,
\begin{equation}
w_{\mathrm{geo}} = 1.00 - 0.20s.
\label{eq:geo_weight}
\end{equation}

At the early stages of training, when predicted boxes are often poorly aligned with the ground truth, geometric terms provide more reliable guidance for correcting large localization errors. As training progresses and predictions become closer to the target boxes, overlap becomes a more informative signal. Increasing the weight of the IoU term therefore encourages the model to focus progressively on improving the overlap between predicted and ground-truth boxes. The geometric terms remain present throughout training to maintain stable spatial alignment and shape consistency.

\subsection{MoEIoU Loss Formulation}\label{subsec:moeiou_formulation}

The final regression objective combines the overlap term~\eqref{eq:log_iou_term}, center-distance term~\eqref{eq:center_term}, and aspect-ratio term~\eqref{eq:aspect_term}:
\begin{equation}
T = [T_{\mathrm{iou}},\; T_{\mathrm{center}},\; T_{\mathrm{aspect}}].
\end{equation}

Rather than summing these components with fixed coefficients, they are aggregated using a weighted LSE formulation:
\begin{equation}
L_{\mathrm{MoEIoU}} =
\frac{1}{\tau}
\log
\left(
\sum_i w_i \exp(\tau T_i)
\right),
\label{eq:proposed_loss2}
\end{equation}
where $T_i$ denotes the $i$-th component, $w_i$ is its corresponding weight, and $\tau > 0$ is a temperature parameter controlling the sharpness of the aggregation.

Using the curriculum-scheduled weights $w_{\mathrm{iou}}$ and $w_{\mathrm{geo}}$ defined in Eq.~\eqref{eq:iou_weight} and Eq.~\eqref{eq:geo_weight}, respectively, $L_{\mathrm{MoEIoU}}$ is expanded as:
\begin{equation}
L_{\mathrm{MoEIoU}} = \frac{1}{\tau}\log \Big(
w_{\mathrm{iou}} e^{\tau T_{\mathrm{iou}}}
+ w_{\mathrm{geo}} e^{\tau T_{\mathrm{center}}}
+ w_{\mathrm{geo}} e^{\tau T_{\mathrm{aspect}}}
\Big).
\label{eq:proposed_loss1}
\end{equation}

This form admits a natural mixture-of-experts interpretation~\cite{Jacobs1991AdaptiveMO}. Each term can be
viewed as an expert focusing on a distinct aspect of localization quality: overlap, positional alignment, and shape consistency. The LSE operator does not merely average these signals; instead, it emphasizes the terms that are currently most informative, while still preserving contribution from the others. In this sense, allowing the optimization process to attend more strongly to whichever regression error dominates at a given moment.

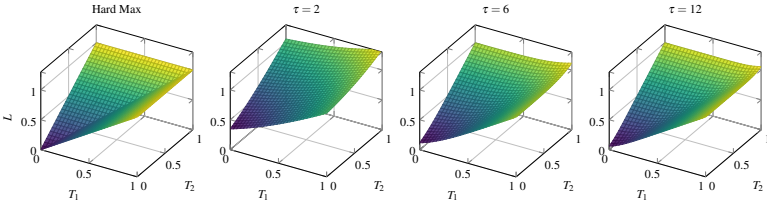
\begin{figure}[H]
\centering
\resizebox{0.8\textwidth}{!}{
\begin{tikzpicture}

% ===================== HARD MAX =====================
\begin{axis}[
    name=plot1,
    width=6cm, height=6cm,
    view={30}{35},
    title={Hard Max},
    xlabel={$T_1$},
    ylabel={$T_2$},
    zlabel={$L$},
    domain=0:1,
    y domain=0:1,
    samples=30,
    colormap/viridis,
    grid=major,
    zmin=0, zmax=1.3
]
\addplot3[surf, opacity=0.9] {max(x,y)};
\end{axis}

% ===================== TAU = 2 =====================
\begin{axis}[
    name=plot2,
    at={(plot1.outer east)},
    anchor=outer west,
    width=6cm, height=6cm,
    view={30}{35},
    title={\textbf{$\tau = 2$}},
    xlabel={$T_1$},
    ylabel={$T_2$},
    domain=0:1,
    y domain=0:1,
    samples=30,
    colormap/viridis,
    grid=major,
    zmin=0, zmax=1.3
]
\addplot3[surf, opacity=0.9] {ln(exp(2*x)+exp(2*y))/2};
\end{axis}

% ===================== TAU = 6 =====================
\begin{axis}[
    name=plot3,
    at={(plot2.outer east)},
    anchor=outer west,
    width=6cm, height=6cm,
    view={30}{35},
    title={\textbf{$\tau = 6$}},
    xlabel={$T_1$},
    ylabel={$T_2$},
    domain=0:1,
    y domain=0:1,
    samples=30,
    colormap/viridis,
    grid=major,
    zmin=0, zmax=1.3
]
\addplot3[surf, opacity=0.9] {ln(exp(6*x)+exp(6*y))/6};
\end{axis}

% ===================== TAU = 12 =====================
\begin{axis}[
    name=plot4,
    at={(plot3.outer east)},
    anchor=outer west,
    width=6cm, height=6cm,
    view={30}{35},
    title={\textbf{$\tau = 12$}},
    xlabel={$T_1$},
    ylabel={$T_2$},
    domain=0:1,
    y domain=0:1,
    samples=30,
    colormap/viridis,
    grid=major,
    zmin=0, zmax=1.3
]
\addplot3[surf, opacity=0.9] {ln(exp(12*x)+exp(12*y))/12};
\end{axis}

\end{tikzpicture}
}
\caption{Comparison of hard maximum and Log-Sum-Exp surfaces over $T_1,T_2 \in [0,1]$.}
\label{fig:lse_tau_comparison}
\end{figure}

For small values of temperature parameter $\tau$, the aggregation behaves similarly to a weighted average, allowing all components to contribute. 
As $\tau$ increases, the aggregation becomes progressively sharper, placing greater emphasis on the component with the largest value and reducing the influence of the others; for large $\tau$, it behaves like a hard maximum.
Figure~\ref{fig:lse_tau_comparison} illustrates an example of this effect for the LSE formulation over two terms. 
The hard maximum forms a sharp ridge along the boundary where the two components are equal, at which the function is not differentiable. 
In contrast, the LSE formulation remains smooth and differentiable across the entire domain.
For lower values of $\tau$, the transition between components is gradual, while for larger values of $\tau$, it becomes increasingly sharp and more closely approximates the hard maximum without introducing non-differentiability.

\subsection{Effect of LSE Aggregation on Gradient Behavior}
\label{subsec:gradient_behavior_lse}

Although the proposed use of LSE aggregation and the conventional weighted sum both combine multiple regression components, they lead to fundamentally different optimization behavior.

Differentiating proposed MoEIoU loss defined in eq.\ref{eq:proposed_loss2} with respect to $T_i$ gives
\begin{equation}
\frac{\partial L_{\mathrm{MoEIoU}}}{\partial T_i}
=
\frac{w_i e^{\tau T_i}}{\sum_j w_j e^{\tau T_j}}.
\end{equation}
This derivative has a normalized softmax-like form where the gradient assigned to each component is determined dynamically by its relative magnitude. Components with larger values receive stronger gradient emphasis, while smaller components still retain nonzero differentiable contributions.

This differs fundamentally from a conventional weighted sum,
\begin{equation}
L_{\mathrm{sum}}=\sum_i w_i T_i,
\quad \text{for which} \quad
\frac{\partial L_{\mathrm{sum}}}{\partial T_i}=w_i.
\end{equation}
Under the weighted-sum formulation, each term contributes according to a fixed coefficient regardless of whether it is currently dominant or negligible. In contrast, LSE adapts the gradient distribution to the current error profile of each sample.

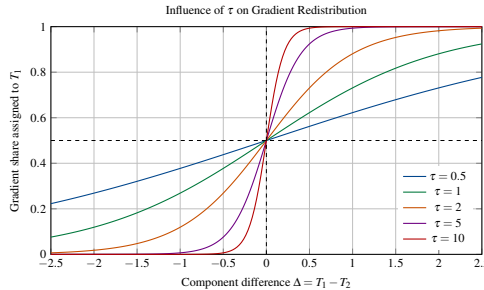
\begin{figure}[htbp]
\centering
\resizebox{0.50\linewidth}{!}{%
\begin{tikzpicture}

% STRONG, HIGH-CONTRAST COLORS
\definecolor{cbRed}{RGB}{180,0,0}
\definecolor{cbBlue}{RGB}{0,70,140}
\definecolor{cbGreen}{RGB}{0,120,60}
\definecolor{cbPurple}{RGB}{110,0,130}
\definecolor{cbOrange}{RGB}{200,80,0}

\begin{axis}[
    width=0.99\columnwidth,
    height=7.5cm,
    xmin=-2.5, xmax=2.5,
    ymin=0, ymax=1,
    xlabel={Component difference $\Delta = T_1 - T_2$},
    ylabel={Gradient share assigned to $T_1$},
    title={Influence of $\tau$ on Gradient Redistribution},
    grid=major,
    legend style={
        at={(0.98,0.02)},
        anchor=south east,
        font=\small,
        draw=none,
        fill=white
    },
    ticklabel style={font=\small},
    label style={font=\small},
    title style={font=\small},
    samples=200,
    domain=-2.5:2.5
]

\addplot[thick, color=cbBlue]   {1/(1 + exp(-0.5*x))};
\addlegendentry{$\tau=0.5$}

\addplot[thick, color=cbGreen]  {1/(1 + exp(-1.0*x))};
\addlegendentry{$\tau=1$}

\addplot[thick, color=cbOrange] {1/(1 + exp(-2.0*x))};
\addlegendentry{$\tau=2$}

\addplot[thick, color=cbPurple] {1/(1 + exp(-5.0*x))};
\addlegendentry{$\tau=5$}

\addplot[thick, color=cbRed]    {1/(1 + exp(-10.0*x))};
\addlegendentry{$\tau=10$}

\addplot[dashed, black] coordinates {(0,0) (0,1)};
\addplot[dashed, black] coordinates {(-2.5,0.5) (2.5,0.5)};

\end{axis}
\end{tikzpicture}%
}
\caption{Gradient redistribution under different temperature values $\tau$.}
\label{fig:tau_gradient_allocation}
\end{figure}

The temperature parameter $\tau$ controls how the gradient is distributed across different components as illustrated in Figure~\ref{fig:tau_gradient_allocation}. It shows how the gradient assigned to $T_1$ varies with the difference between the two terms, defined as $\Delta = T_1 - T_2$. 
When $\Delta = 0$, both terms receive equal gradient contribution.  
For small $\tau$, the gradients are assigned more uniformly, indicating that it is shared smoothly between the two components. As $\tau$ increases, the transition becomes steeper, so even a small difference between $T_1$ and $T_2$ leads to most of the gradient being assigned to the larger term. Thus, $\tau$ determines the transition from balanced optimization across all components to focusing primarily on the dominant error term.

For bounding box regression, some predictions may primarily suffer from poor overlap, whereas others may be affected more strongly by center displacement or shape mismatch. The LSE formulation therefore allows the optimization process to respond automatically to the most critical localization error in each case, without requiring manually designed switching rules or case-specific heuristics.

%-------------------------------------------------------------------------
\section{Simulation Experiment}
\label{sec:simulation_experiment}

We evaluate bounding-box regression losses using a controlled simulation experiment to compare how different losses guide the optimization of a predicted box toward the ground-truth box. In object detection models, the regression loss influences performance through backpropagation and gradient descent based parameter updates. As a result, it is difficult to isolate the individual effect of the loss function from the overall optimization process and final detection performance.

To focus only on the behavior of the regression loss, instead of running through the overall optimization process of training a detector, we directly update the predicted bounding box using the gradient of the selected loss with respect to a fixed ground-truth box. Starting from an initial predicted box, the box parameters are iteratively updated until they move closer to the ground-truth box.

\subsection{Simulation Setup}
\label{subsec:simulation_setup}
We generate 10 million synthetic box pairs, where each sample consists of a ground-truth
box and an initial predicted box represented in center-size form $(c_x,c_y,w,h)$. The
synthetic pairs are designed to mimic common localization errors produced by real-world
object detectors. The
ground-truth box centers are sampled from $[7,13]\times[7,13]$, with box area sampled
log-uniformly from $[0.1,4.0]$ and aspect ratio sampled log-uniformly from $[0.2,5.0]$.
The initial predicted boxes are then sampled to cover eight distinct kinds of localization
errors:
far non-overlap, where the prediction is shifted by roughly $1.5$--$4.0$ times the box
size; near non-overlap, where the prediction is close to the target but has little or no
overlap; low, medium, and high-overlap cases with initial IoU ranges $[0.05,0.30]$,
$[0.30,0.60]$, and $[0.60,0.95]$, respectively; center-shape mismatch, where the center
is correct but width and height are inaccurate; shifted boxes, where the shape is correct
but the center is displaced; and nested boxes, where the prediction lies inside the
ground truth. The distribution is intentionally biased toward difficult regression cases,
with approximately 40\% far non-overlapping predictions and the remaining 60\% distributed
across the other seven categories.

\begin{equation}
\scalebox{0.82}{$
B_n^{(t)} =
B_n^{(t-1)}
-
\eta_t
\bigl(2-\mathrm{IoU}(B_n^{(t-1)},G_n)\bigr)
\nabla_{B_n^{(t-1)}}
\mathcal{L}(B_n^{(t-1)},G_n)
$}.
\label{eq:simulation_update}
\end{equation}

We adopt the DIoU/CIoU simulation algorithm~\cite{Zheng2020DIoU, ZhengCIoU} for bounding box optimization. 
For a regression loss $\mathcal{L}$, each predicted box $B_n^{(t)}$ is updated toward its
corresponding ground-truth box $G_n$ using the gradient-descent update in
Eq.~\ref{eq:simulation_update}. The factor $(2-\mathrm{IoU})$ increases the update
magnitude for poorly overlapping boxes and reduces it as the prediction approaches the
target. The learning rate ($\eta_t$) uses the same three-stage schedule with
$\eta_t=0.1$ for the first 80\% of iterations, $\eta_t=0.01$ for the next 10\%, and
$\eta_t=0.001$ for the final 10\%. Box widths and heights are clamped after each update to
remain positive.

\subsection{Evaluation}
\label{subsec:evaluation}

To evaluate the optimization behavior of different regression losses, we track the mean IoU across all simulated cases during the optimization process. At each iteration, the IoU between the predicted box and the corresponding ground-truth box is computed, and the average IoU over all box pairs is recorded. 

\begin{figure}[H]
\centering

\begin{minipage}[t]{0.70\textwidth}
\centering
\includegraphics[width=\linewidth]{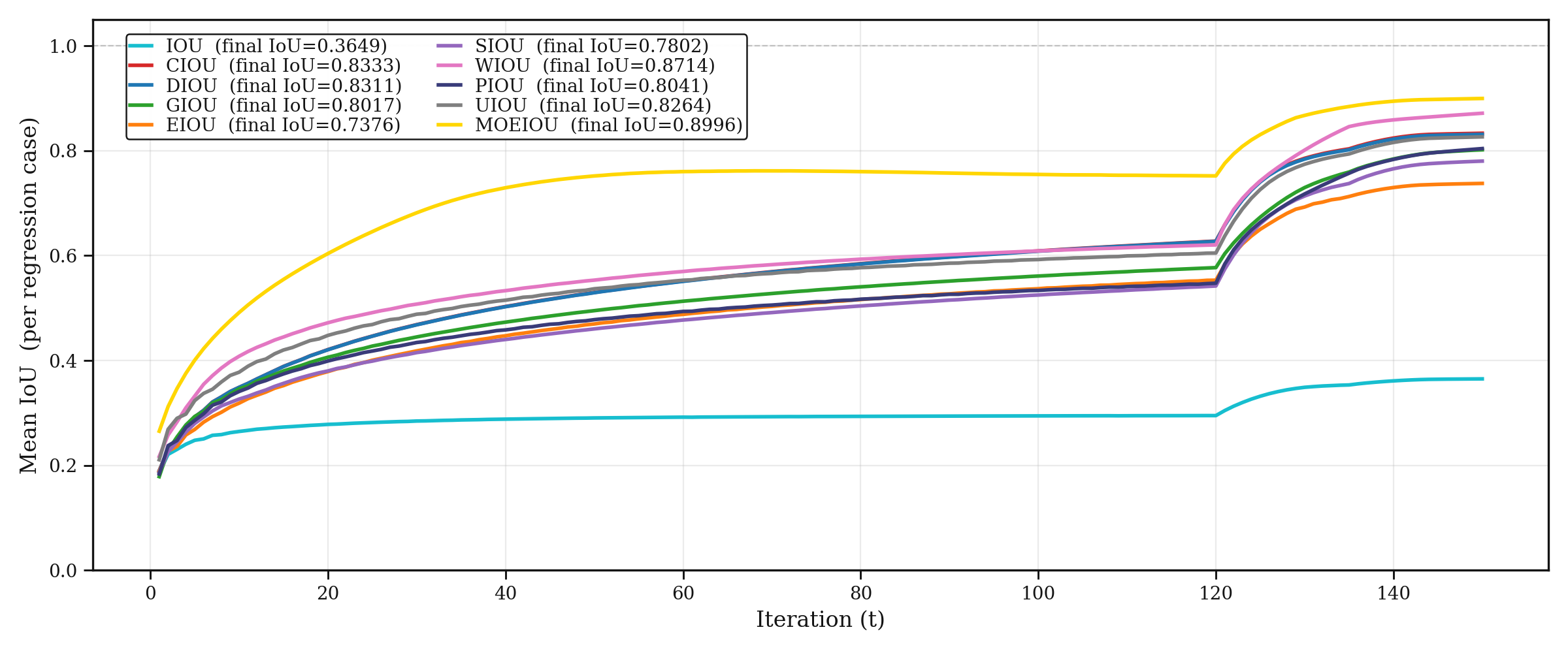}
\vspace{1mm}

{\small (a) Mean IoU over iterations}
\end{minipage}
\hfill
\begin{minipage}[t]{0.28\textwidth}
\centering
\includegraphics[width=\linewidth]{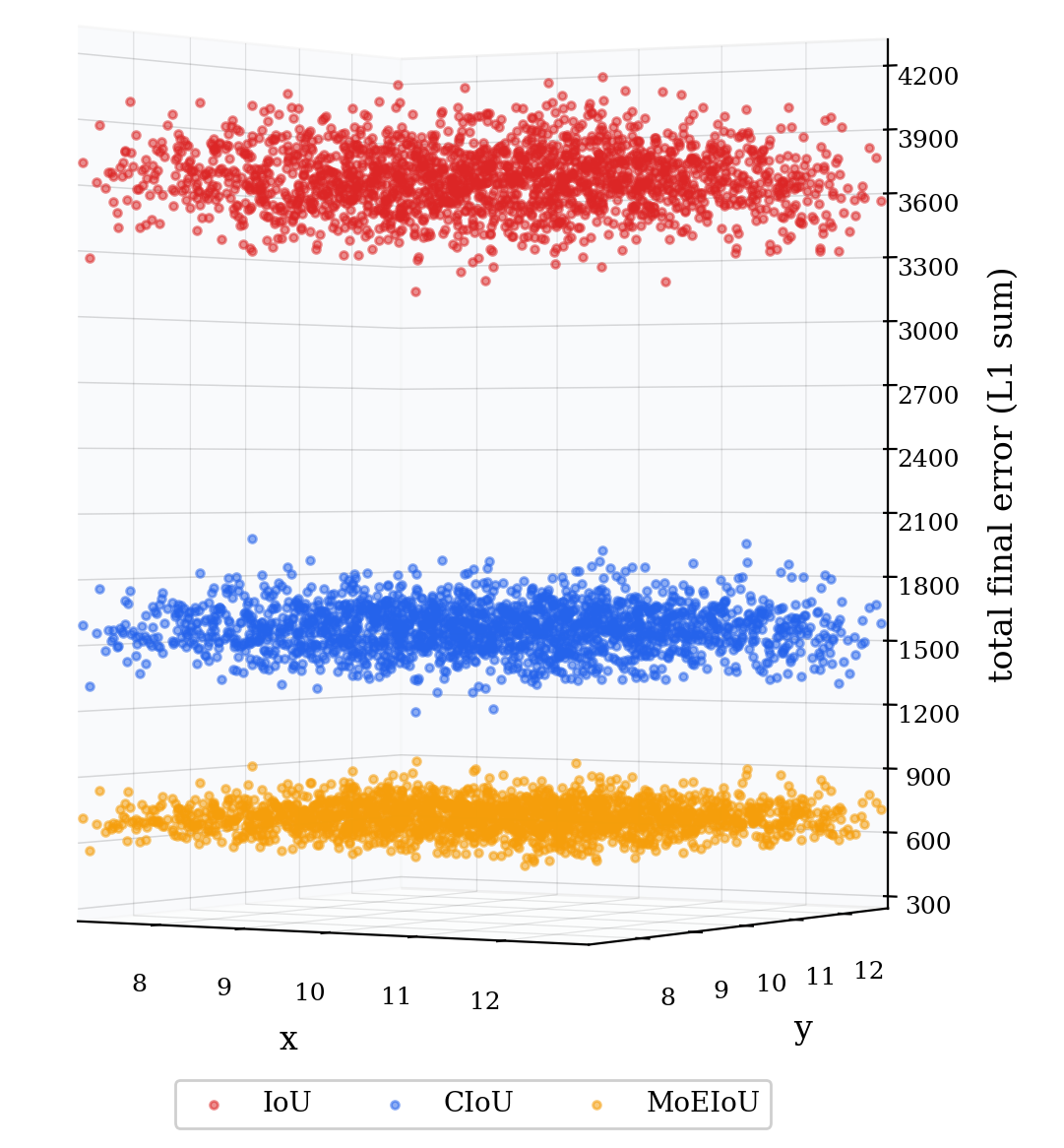}
\vspace{1mm}

{\small (b) Final regression error}
\end{minipage}

\caption{
Simulation-based comparison of bounding-box regression losses.
(a) Mean IoU over optimization iterations.
(b) Final regression error produced by IoU, CIoU, and MoEIoU.
}
\label{fig:simulation_iou_error_sidebyside}
\end{figure}

Figure~\ref{fig:simulation_iou_error_sidebyside} (a) shows the mean IoU over 150 optimization iterations for all compared regression losses.  After 150 iterations, MoEIoU reaches the highest final mean IoU of 0.8996. The next best result is obtained by WIoU~\cite{Tong2023WIoU} with a final mean IoU of 0.8714. An important observation is that the advantage of MoEIoU is not limited to the final mean IoU. Among all compared methods, the proposed MoEIoU's curve remains above all other losses throughout the optimization process. This indicates that MoEIoU provides a stronger optimization signal across different stages of box refinement, leading to both faster convergence and better final localization quality.

After optimization, for each anchor center $(x,y)$, the final regression error is computed
as the L1 distance between the predicted box parameters and the ground-truth box parameters. Figure~\ref{fig:simulation_iou_error_sidebyside} (b) visualizes this final regression error across all anchor locations. It depicts a 3D scatter plot where each point corresponds to one optimization run starting from a specific location $(x,y)$. The height of each point represents the total final regression error after $T$ iterations. 
The basic IoU loss $(1-\text{IoU})$, which is one of the earliest bounding-box regression losses~\cite{Yu2016UnitBox}, produces the largest errors and the widest spread across spatial locations. CIoU~\cite{ZhengCIoU}, which is used in modern detectors, substantially reduces both the magnitude and variability of the regression error. The proposed MoEIoU further improves this behavior, producing the lowest errors and the most concentrated distribution. This indicates more accurate and stable bounding-box regression during optimization.

%-------------------------------------------------------------------------

%-------------------------------------------------------------------------
\section{Results}
\label{sec:results}

\subsection{Experimental Setup}
All experiments use the Ultralytics implementation~\cite{Jocher_Ultralytics_YOLO_2023}. 
For a fair comparison, only the bounding-box regression loss is changed; the detector 
architecture, data processing, and training configuration are kept fixed within each setting. 
Unless otherwise stated, models are trained at $640\times640$ resolution with batch size 16 
using SGD with learning rate 0.01, momentum 0.9, and weight decay $5\times10^{-4}$. 
We fix the random seed to 42 for Python, NumPy, and PyTorch, enable deterministic CuDNN 
behavior, and run experiments on an NVIDIA RTX A6000 GPU. For MoEIoU, we choose the temperature parameter $\tau$ through short trial runs before full
training. We first train for a few epochs using $\tau \in \{1,4,8,10,12\}$ and select the
value with the best validation result. We then check nearby values for a few more epochs if
needed. We did not observe
additional gains for $\tau>12$ on either benchmark datasets or simulation experiments.

\subsection{Datasets}
\label{subsec:datasets}

We evaluate MoEIoU on three object-detection benchmarks: PASCAL VOC~\cite{PascalVOC}, 
HRIPCB~\cite{HRIPCB}, and MS COCO~\cite{Lin2014COCO}. PASCAL VOC is used to evaluate 
general object detection across 20 everyday object categories with variations in scale, pose, 
occlusion, and background clutter. We train using the union of the VOC2007 and VOC2012 
train/val splits and evaluate on the VOC2007 test split.

HRIPCB~\cite{HRIPCB} is used to evaluate small-defect localization in industrial PCB inspection. It contains 
1,386 annotated PCB images with six defect categories: missing hole, mouse bite, open circuit, 
short, spur, and spurious copper. These defects are often small and visually similar to 
surrounding circuit patterns, making precise localization challenging~\cite{Smalldefectdetection}.

For MS COCO, we use the curated 20\% subset of the train2017 split introduced by prior 
work~\cite{HoughNet}. This subset contains approximately 25K images while preserving the 
class distribution, object-scale statistics, and difficulty of the full train2017 set. COCO 
contains complex everyday scenes with 80 object categories and large variation in object 
scale, viewpoint, occlusion, and scene composition.

\subsection{Benchmark Results}
\label{subsec:benchmark_results}

The main benchmark results are reported in Tables~\ref{tab:voc_results}--\ref{tab:coco_results}. 
Across the three datasets, only the bounding-box regression loss is changed, while the model 
architecture and training configuration are kept fixed within each setting. 

\begin{table*}[!ht]
  \centering
  \caption{Comparison of different bounding-box regression losses on the PASCAL VOC.}
  \resizebox{0.7\textwidth}{!}{%
    \begin{tabular}{lcccccc}
      \toprule
      \multirow{2}{*}{\textbf{\makecell[l]{Loss}}} &
      \multicolumn{3}{c}{\textbf{YOLOv12}} &
      \multicolumn{3}{c}{\textbf{YOLO26}} \\
      \cmidrule(lr){2-4}
      \cmidrule(lr){5-7}
      &
      \textbf{\makecell[c]{mAP$_{50}$}} &
      \textbf{\makecell[c]{mAP$_{75}$}} &
      \textbf{\makecell[c]{mAP$_{50:95}$}} &
      \textbf{\makecell[c]{mAP$_{50}$}} &
      \textbf{\makecell[c]{mAP$_{75}$}} &
      \textbf{\makecell[c]{mAP$_{50:95}$}} \\
      \midrule

      \makecell[l]{DIoU (2019)} &
      \makecell[c]{75.4} & \makecell[c]{59.6} & \makecell[c]{54.3} &
      \makecell[c]{72.7} & \makecell[c]{57.0} & \makecell[c]{51.8} \\

      \makecell[l]{GIoU (2019)} &
      \makecell[c]{75.1} & \makecell[c]{60.1} & \makecell[c]{54.4} &
      \makecell[c]{72.8} & \makecell[c]{57.0} & \makecell[c]{51.9} \\

      \makecell[l]{CIoU (2020)} &
      \makecell[c]{74.8} & \makecell[c]{58.8} & \makecell[c]{53.9} &
      \makecell[c]{72.3} & \makecell[c]{56.9} & \makecell[c]{51.4} \\

      \makecell[l]{EIoU (2021)} &
      \makecell[c]{74.5} & \makecell[c]{58.6} & \makecell[c]{53.2} &
      \makecell[c]{72.4} & \makecell[c]{56.3} & \makecell[c]{51.1} \\

      \makecell[l]{SIoU (2022)} &
      \makecell[c]{75.4} & \makecell[c]{59.7} & \makecell[c]{54.3} &
      \makecell[c]{73.4} & \makecell[c]{56.9} & \makecell[c]{52.1} \\

      \makecell[l]{WIoU (2023)} &
      \makecell[c]{75.2} & \makecell[c]{59.4} & \makecell[c]{53.9} &
      \makecell[c]{72.9} & \makecell[c]{57.8} & \makecell[c]{51.6} \\

      \makecell[l]{PIoU (2024)} &
      \makecell[c]{75.4} & \makecell[c]{59.5} & \makecell[c]{54.2} &
      \makecell[c]{72.6} & \makecell[c]{56.8} & \makecell[c]{51.4} \\

      \makecell[l]{UIoU (2024)} &
      \makecell[c]{75.3} & \makecell[c]{59.4} & \makecell[c]{54.2} &
      \makecell[c]{72.6} & \makecell[c]{56.5} & \makecell[c]{51.4} \\

      \makecell[l]{\textbf{MoEIoU}} &
      \makecell[c]{\textbf{76.2}} & \makecell[c]{\textbf{61.0}} & \makecell[c]{\textbf{55.0}} &
      \makecell[c]{\textbf{74.1}} & \makecell[c]{\textbf{57.9}} & \makecell[c]{\textbf{52.4}} \\

      \bottomrule
    \end{tabular}
  }

  {
  \scriptsize
  \begin{minipage}{0.8\textwidth}
  \begin{flushleft}
  \textbf{Note:} Best results in each column are shown in bold. IoU denotes Intersection over Union, and mAP denotes mean Average Precision.
  \end{flushleft}
  \end{minipage}
  }

  \label{tab:voc_results}
\end{table*}

\begin{table*}[!ht]
  \centering
  \caption{Comparison of different bounding-box regression losses on the HRIPCB dataset.}
  \resizebox{0.7\textwidth}{!}{%
    \begin{tabular}{lcccccc}
      \toprule
      \multirow{2}{*}{\textbf{\makecell[l]{Loss}}} &
      \multicolumn{3}{c}{\textbf{YOLOv12}} &
      \multicolumn{3}{c}{\textbf{YOLO26}} \\
      \cmidrule(lr){2-4}
      \cmidrule(lr){5-7}
      &
      \textbf{\makecell[c]{mAP$_{50}$}} &
      \textbf{\makecell[c]{mAP$_{75}$}} &
      \textbf{\makecell[c]{mAP$_{50:95}$}} &
      \textbf{\makecell[c]{mAP$_{50}$}} &
      \textbf{\makecell[c]{mAP$_{75}$}} &
      \textbf{\makecell[c]{mAP$_{50:95}$}} \\
      \midrule

      \makecell[l]{DIoU (2019)} &
      \makecell[c]{98.0} & \makecell[c]{61.8} & \makecell[c]{58.4} &
      \makecell[c]{98.0} & \makecell[c]{64.4} & \makecell[c]{59.0} \\

      \makecell[l]{GIoU (2019)} &
      \makecell[c]{97.2} & \makecell[c]{61.2} & \makecell[c]{57.8} &
      \makecell[c]{97.7} & \makecell[c]{64.2} & \makecell[c]{58.9} \\

      \makecell[l]{CIoU (2020)} &
      \makecell[c]{98.1} & \makecell[c]{62.9} & \makecell[c]{58.4} &
      \makecell[c]{97.7} & \makecell[c]{62.7} & \makecell[c]{59.2} \\

      \makecell[l]{EIoU (2021)} &
      \makecell[c]{97.9} & \makecell[c]{60.4} & \makecell[c]{58.1} &
      \makecell[c]{97.9} & \makecell[c]{64.4} & \makecell[c]{58.3} \\

      \makecell[l]{SIoU (2022)} &
      \makecell[c]{97.7} & \makecell[c]{63.2} & \makecell[c]{58.5} &
      \makecell[c]{97.7} & \makecell[c]{64.9} & \makecell[c]{59.0} \\

      \makecell[l]{WIoU (2023)} &
      \makecell[c]{97.6} & \makecell[c]{61.9} & \makecell[c]{58.0} &
      \makecell[c]{97.9} & \makecell[c]{65.9} & \makecell[c]{59.2} \\

      \makecell[l]{PIoU (2024)} &
      \makecell[c]{97.9} & \makecell[c]{59.5} & \makecell[c]{58.1} &
      \makecell[c]{98.0} & \makecell[c]{64.8} & \makecell[c]{59.1} \\

      \makecell[l]{UIoU (2024)} &
      \makecell[c]{98.1} & \makecell[c]{61.9} & \makecell[c]{58.1} &
      \makecell[c]{97.9} & \makecell[c]{65.5} & \makecell[c]{58.6} \\

      \makecell[l]{\textbf{MoEIoU}} &
      \makecell[c]{\textbf{98.6}} & \makecell[c]{\textbf{66.9}} & \makecell[c]{\textbf{59.4}} &
      \makecell[c]{\textbf{98.6}} & \makecell[c]{\textbf{67.2}} & \makecell[c]{\textbf{59.7}} \\

      \bottomrule
    \end{tabular}
  }

  {
  \scriptsize
  \begin{minipage}{0.8\textwidth}
  \begin{flushleft}
  \textbf{Note:} Best results in each column are shown in bold. IoU denotes Intersection over Union, and mAP denotes mean Average Precision.
  \end{flushleft}
  \end{minipage}
  }

  \label{tab:hripcb_results}
\end{table*}

\begin{table*}[!ht]
  \centering
  \caption{Comparison of different bounding-box regression losses on the MS COCO dataset. }
  \resizebox{0.7\textwidth}{!}{%
    \begin{tabular}{lcccccc}
      \toprule
      \multirow{2}{*}{\textbf{\makecell[l]{Loss}}} &
      \multicolumn{3}{c}{\textbf{YOLOv12}} &
      \multicolumn{3}{c}{\textbf{YOLO26}} \\
      \cmidrule(lr){2-4}
      \cmidrule(lr){5-7}
      &
      \textbf{\makecell[c]{mAP$_{50}$}} &
      \textbf{\makecell[c]{mAP$_{75}$}} &
      \textbf{\makecell[c]{mAP$_{50:95}$}} &
      \textbf{\makecell[c]{mAP$_{50}$}} &
      \textbf{\makecell[c]{mAP$_{75}$}} &
      \textbf{\makecell[c]{mAP$_{50:95}$}} \\
      \midrule

      \makecell[l]{DIoU (2019)} &
      \makecell[c]{47.1} & \makecell[c]{35.1} & \makecell[c]{32.5} &
      \makecell[c]{45.8} & \makecell[c]{34.1} & \makecell[c]{31.7} \\

      \makecell[l]{GIoU (2019)} &
      \makecell[c]{46.6} & \makecell[c]{34.8} & \makecell[c]{32.1} &
      \makecell[c]{46.3} & \makecell[c]{34.8} & \makecell[c]{32.1} \\

      \makecell[l]{CIoU (2020)} &
      \makecell[c]{47.5} & \makecell[c]{35.1} & \makecell[c]{32.7} &
      \makecell[c]{46.4} & \makecell[c]{34.7} & \makecell[c]{32.2} \\

      \makecell[l]{EIoU (2021)} &
      \makecell[c]{46.9} & \makecell[c]{34.9} & \makecell[c]{32.3} &
      \makecell[c]{46.1} & \makecell[c]{34.2} & \makecell[c]{31.7} \\

      \makecell[l]{SIoU (2022)} &
      \makecell[c]{46.9} & \makecell[c]{34.9} & \makecell[c]{32.3} &
      \makecell[c]{46.2} & \makecell[c]{34.7} & \makecell[c]{32.0} \\

      \makecell[l]{WIoU (2023)} &
      \makecell[c]{47.7} & \makecell[c]{35.4} & \makecell[c]{32.9} &
      \makecell[c]{46.4} & \makecell[c]{34.7} & \makecell[c]{31.9} \\

      \makecell[l]{PIoU (2024)} &
      \makecell[c]{47.3} & \makecell[c]{34.9} & \makecell[c]{32.6} &
      \makecell[c]{46.2} & \makecell[c]{34.7} & \makecell[c]{32.0} \\

      \makecell[l]{UIoU (2024)} &
      \makecell[c]{47.1} & \makecell[c]{35.0} & \makecell[c]{32.5} &
      \makecell[c]{46.0} & \makecell[c]{34.3} & \makecell[c]{31.8} \\

      \makecell[l]{\textbf{MoEIoU}} &
      \makecell[c]{\textbf{48.6}} & \makecell[c]{\textbf{35.7}} & \makecell[c]{\textbf{33.1}} &
      \makecell[c]{\textbf{46.6}} & \makecell[c]{\textbf{34.9}} & \makecell[c]{\textbf{32.3}} \\

      \bottomrule
    \end{tabular}
  }

  {
  \scriptsize
  \begin{minipage}{0.8\textwidth}
  \begin{flushleft}
  \textbf{Note:} Best results in each column are shown in bold. IoU denotes Intersection over Union, and mAP denotes mean Average Precision.
  \end{flushleft}
  \end{minipage}
  }

  \label{tab:coco_results}
\end{table*}

\begin{figure}[H]
\centering
\resizebox{\textwidth}{!}{%
\begin{minipage}{\textwidth}
\centering

\includegraphics[width=\textwidth]{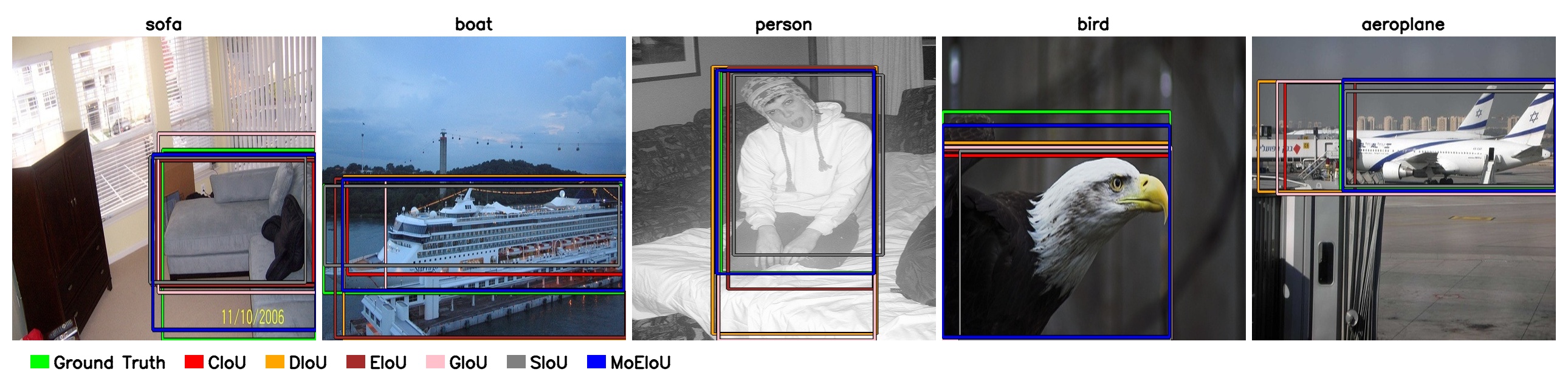}

\end{minipage}%
}

\caption{Comparison of bounding box predictions produced by \textbf{YOLOv12}~\cite{tian2025yolov12attentioncentricrealtimeobject} trained with different IoU-based loss functions on the PASCAL VOC test set~\cite{PascalVOC}}.
\label{fig:stacked_examples_yolo12}
\end{figure}

Across the three benchmarks, MoEIoU achieves the best mAP$_{50:95}$ for both YOLOv12 
and YOLO26. On PASCAL VOC, it improves YOLOv12~\cite{tian2025yolov12attentioncentricrealtimeobject} 
mAP$_{50:95}$ to 55.0 and YOLO26~\cite{yolo26_ultralytics} to 52.4. 
On HRIPCB, MoEIoU gives the strongest strict-localization results. On MS COCO, MoEIoU is best across all reported 
metrics for both detectors. The qualitative YOLOv12 results in Fig.~\ref{fig:stacked_examples_yolo12} 
further show that MoEIoU produces tighter and more accurate bounding boxes compared with 
other IoU-based losses. These results indicate that the proposed loss improves 
localization quality across both general-object and small-defect detection settings.

\subsection{Effect of LSE Aggregation}
\label{subsec:lse_results}

\begin{figure}[H]
\centering

\includegraphics[width=0.9\textwidth]{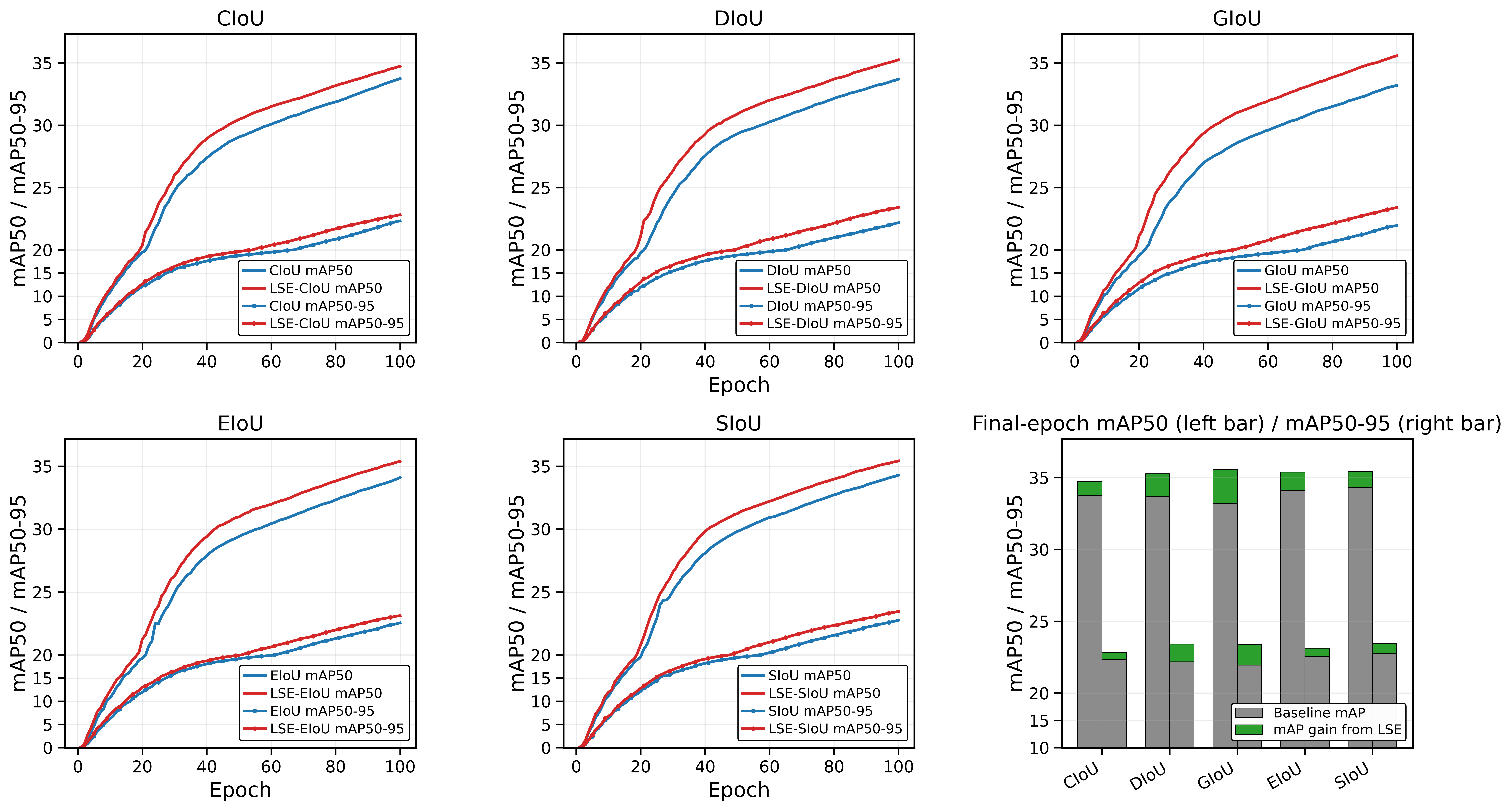}
\caption{
Comparison of validation mAP curves and final-epoch mAP gains. 
}
\label{fig:lse_iou_variants_coco}
\end{figure}

To evaluate the effect of the aggregation strategy, we replace the additive formulation of 
CIoU, DIoU, GIoU, EIoU, and SIoU with LSE aggregation while keeping their original 
geometric terms unchanged. These losses are selected because they contain multiple geometric terms, making them suitable for LSE-based aggregation, which adaptively emphasizes the dominant localization error among participating terms. WIoU is excluded since it mainly uses a single geometric term, while PIoU and UIoU already include their own weighting mechanisms, making LSE redundant. All variants are trained on MS COCO using the same YOLOv12n 
configuration for 100 epochs. Table~\ref{tab:loss_formulas_results} shows that LSE aggregation improves every evaluated IoU-based 
loss. The gains are consistent on mAP$_{50:95}$, with relative improvements of +2.23\% for 
CIoU, +5.57\% for DIoU, +6.62\% for GIoU, +2.52\% for EIoU, and +3.09\% for SIoU. 
These results demonstrate the effectiveness of LSE as an adaptive aggregation mechanism. Training curves for these variants are provided in Fig~\ref{fig:lse_iou_variants_coco}.

\newcommand{\termcell}[1]{\makecell[l]{\scriptsize #1}}

\begin{table}[!ht]
  \centering
  \caption{IoU-based loss formulations, their LSE counterparts, and performance comparison.}
  \resizebox{0.8\textwidth}{!}{%
  \begin{tabular}{l l c c c c}
    \toprule
    \textbf{\makecell[l]{Loss}} &
    \textbf{\makecell[l]{Formulation}} &
    \textbf{\makecell[c]{mAP$_{50}$}} &
    \textbf{\makecell[c]{mAP$_{65}$}} &
    \textbf{\makecell[c]{mAP$_{75}$}} &
    \textbf{\makecell[c]{mAP$_{50:95}$}} \\
    \midrule

      \makecell[l]{CIoU} &
    \termcell{
    $T_{\mathrm{iou}}=1-\mathrm{IoU}$,\quad
    $T_{\mathrm{center}}=\frac{\rho^2}{c^2}$,\quad
    $T_{\mathrm{aspect}}=\alpha v$
    } &
    \makecell[c]{33.75} & \makecell[c]{29.01} & \makecell[c]{23.68} & \makecell[c]{22.32} \\

    \makecell[l]{LSE-CIoU} &
    \termcell{
    $L_{\mathrm{LSE\text{-}CIoU}}=
    \mathrm{LSE}(T_{\mathrm{iou}},T_{\mathrm{center}},T_{\mathrm{aspect}})$
    } &
    \makecell[c]{35.00} & \makecell[c]{29.87} & \makecell[c]{24.36} & \makecell[c]{22.72} \\

    \makecell[l]{$\Delta$ (\%)} &
    \termcell{Relative gain over CIoU} &
    \makecell[c]{+3.70} & \makecell[c]{+2.96} & \makecell[c]{+2.87} & \makecell[c]{+1.79} \\
    
    \midrule

    \makecell[l]{DIoU} &
    \termcell{
    $T_{\mathrm{iou}}=1-\mathrm{IoU}$,\quad
    $T_{\mathrm{center}}=\frac{\rho^2}{c^2}$
    } &
    \makecell[c]{33.70} & \makecell[c]{29.12} & \makecell[c]{23.61} & \makecell[c]{22.18} \\

    \makecell[l]{LSE-DIoU} &
    \termcell{
    $L_{\mathrm{LSE\text{-}DIoU}}=
    \mathrm{LSE}(T_{\mathrm{iou}},T_{\mathrm{center}})$
    } &
    \makecell[c]{35.26} & \makecell[c]{30.66} & \makecell[c]{25.25} & \makecell[c]{23.41} \\

    \makecell[l]{$\Delta$ (\%)} &
    \termcell{ Relative gain over DIoU} &
    \makecell[c]{+4.63} & \makecell[c]{+5.29} & \makecell[c]{+6.94} & \makecell[c]{+5.57} \\

    \midrule

    \makecell[l]{GIoU} &
    \termcell{
    $T_{\mathrm{iou}}=1-\mathrm{IoU}$,\quad
    $T_{\mathrm{enclosure}}=\frac{|C \setminus (A \cup B)|}{|C|}$
    } &
    \makecell[c]{33.20} & \makecell[c]{28.72} & \makecell[c]{23.42} & \makecell[c]{21.95} \\

    \makecell[l]{LSE-GIoU} &
    \termcell{
    $L_{\mathrm{LSE\text{-}GIoU}}=
    \mathrm{LSE}(T_{\mathrm{iou}},T_{\mathrm{enclosure}})$
    } &
    \makecell[c]{35.57} & \makecell[c]{30.86} & \makecell[c]{24.91} & \makecell[c]{23.40} \\

    \makecell[l]{$\Delta$ (\%)} &
    \termcell{ Relative gain over GIoU} &
    \makecell[c]{+7.16} & \makecell[c]{+7.43} & \makecell[c]{+6.37} & \makecell[c]{+6.62} \\

    \midrule

    \makecell[l]{EIoU} &
    \termcell{
    $T_{\mathrm{iou}}=1-\mathrm{IoU}$,\quad
    $T_{\mathrm{center}}=\frac{\rho^2}{c^2}$,\quad 
    $T_{\mathrm{shape}}=
    \frac{(w-w^{gt})^2}{w_c^2}
    +
    \frac{(h-h^{gt})^2}{h_c^2}$
    } &
    \makecell[c]{34.10} & \makecell[c]{29.55} & \makecell[c]{24.03} & \makecell[c]{22.55} \\

    \makecell[l]{LSE-EIoU} &
    \termcell{
    $L_{\mathrm{LSE\text{-}EIoU}}=
    \mathrm{LSE}(T_{\mathrm{iou}},T_{\mathrm{center}},T_{\mathrm{shape}})$
    } &
    \makecell[c]{35.38} & \makecell[c]{30.56} & \makecell[c]{24.60} & \makecell[c]{23.12} \\

    \makecell[l]{$\Delta$ (\%)} &
    \termcell{ Relative gain over EIoU} &
    \makecell[c]{+3.75} & \makecell[c]{+3.43} & \makecell[c]{+2.37} & \makecell[c]{+2.52} \\

    \midrule

    \makecell[l]{SIoU} &
    \termcell{
    $T_{\mathrm{iou}}=1-\mathrm{IoU}$,\quad
    $T_{\mathrm{dist}}=\frac{\Delta}{2}$,\quad
    $T_{\mathrm{shape}}=\frac{\Omega}{2}$
    } &
    \makecell[c]{34.29} & \makecell[c]{29.80} & \makecell[c]{24.57} & \makecell[c]{22.75} \\

    \makecell[l]{LSE-SIoU} &
    \termcell{
    $L_{\mathrm{LSE\text{-}SIoU}}=
    \mathrm{LSE}(T_{\mathrm{iou}},T_{\mathrm{dist}},T_{\mathrm{shape}})$
    } &
    \makecell[c]{35.42} & \makecell[c]{30.73} & \makecell[c]{25.33} & \makecell[c]{23.46} \\

    \makecell[l]{$\Delta$ (\%)} &
    \termcell{ Relative gain over SIoU} &
    \makecell[c]{+3.28} & \makecell[c]{+3.15} & \makecell[c]{+3.10} & \makecell[c]{+3.09} \\

    \bottomrule
  \end{tabular}
  }

  {
  \scriptsize
  \begin{minipage}{0.95\textwidth}
  \begin{flushleft}
  \textbf{Note:}
  For each baseline loss, the loss is the fixed summation of the listed penalty terms.
  The corresponding LSE variant uses the same terms but replaces fixed summation with
  $\mathrm{LSE}(T_1,\dots,T_n)=\frac{1}{\tau}\log\sum_i \exp(\tau T_i)$.
  Here, $\rho$ is the center distance, $c$ is the enclosing-box diagonal,
  $A$ and $B$ are the predicted and ground-truth boxes, $C$ is their smallest enclosing box,
  $w_c,h_c$ are enclosing-box dimensions, $v$ and $\alpha$ are the CIoU aspect-ratio term and balancing coefficient,
  and $\Delta,\Omega$ denote the SIoU distance and shape costs.
  \end{flushleft}
  \end{minipage}
  }

  \label{tab:loss_formulas_results}
\end{table}

%-------------------------------------------------------------------------

\section{Ablation Study}\label{sec:ablation_main}
We conduct ablation studies using the YOLOv12n model on the curated MS COCO dataset~\cite{HoughNet} to evaluate the contribution of each design component in MoEIoU. 

\begin{table} [H]
  \centering
  \caption{Effect of individual MoEIoU components}
  \resizebox{0.7\textwidth}{!}{%
    \begin{tabular}{ccccccc}
      \toprule
      \textbf{\makecell[l]{CIoU Loss}} &
      \textbf{\makecell[l]{LogSumExp}} &
      \textbf{\makecell[l]{Log-IoU}} &
      \textbf{\makecell[l]{Curriculum}} &
      \textbf{\makecell[l]{mAP$_{50}$}} &
      \textbf{\makecell[l]{mAP$_{75}$}} &
      \textbf{\makecell[l]{mAP$_{50:95}$}} \\
      \midrule

      \makecell[l]{\checkmark} & \makecell[l]{$\times$} & \makecell[l]{$\times$} & \makecell[l]{$\times$} &
      \makecell[l]{33.7} & \makecell[l]{23.7} & \makecell[l]{22.3} \\

      \makecell[l]{\checkmark} & \makecell[l]{\checkmark} & \makecell[l]{$\times$} & \makecell[l]{$\times$} &
      \makecell[l]{35.0} & \makecell[l]{24.1} & \makecell[l]{22.7} \\

      \makecell[l]{\checkmark} & \makecell[l]{\checkmark} & \makecell[l]{\checkmark} & \makecell[l]{$\times$} &
      \makecell[l]{35.0} & \makecell[l]{24.4} & \makecell[l]{22.8} \\

      \makecell[l]{\checkmark} & \makecell[l]{\checkmark} & \makecell[l]{\checkmark} & \makecell[l]{\checkmark} &
      \makecell[l]{\textbf{35.0}} & \makecell[l]{\textbf{24.7}} & \makecell[l]{\textbf{23.0}} \\

      \bottomrule
    \end{tabular}
  }
  \label{tab:ablation_coco}
\end{table}

\noindent \textbf{Effect of individual MoEIoU components.} We use standard CIoU as the baseline and incrementally introduce the MoEIoU components, keeping all training settings fixed for 100 epochs with a patience of 10. As reported in Table~\ref{tab:ablation_coco}, replacing the additive CIoU formulation with LSE aggregation improves mAP$_{50}$ from 33.7 to 35.0, mAP$_{75}$ from 23.7 to 24.1, and mAP$_{50:95}$ from 22.3 to 22.7, showing the benefit of adaptively emphasizing the dominant localization error. Adding the Log-IoU overlap term further improves mAP at higher thresholds, increasing mAP$_{75}$ to 24.4 and mAP$_{50:95}$ to 22.8. Finally, introducing curriculum-based weighting gives the best overall result, further raising mAP$_{75}$ to 24.7 and mAP$_{50:95}$ to 23.0.

%We conduct ablation studies on the curated MS COCO dataset~\cite{HoughNet}. To evaluate the contribution of each design component in MoEIoU. All experiments are performed using the YOLOv12n model. We begin with the standard CIoU loss as the baseline and progressively transform it into the proposed formulation. Specifically, we first replace the additive aggregation with the LSE formulation, then introduce the logarithmic overlap term (Log-IoU), and finally incorporate the curriculum-based weighting scheme. This step-by-step progression allows us to isolate the impact of each modification on detection performance. Unless otherwise specified, all experiments are conducted for 100 epochs with a patience of 10 under identical training settings. 

\begin{table*}[!ht]
  \centering
  \caption{Ablation study on geometric term rescaling under LSE aggregation.}
  \resizebox{0.7\textwidth}{!}{%
    \begin{tabular}{lllccc}
      \toprule
      \textbf{\makecell[l]{Formulation}} &
      \textbf{\makecell[l]{Center Term}} &
      \textbf{\makecell[l]{Aspect Term}} &
      \textbf{\makecell[l]{mAP$_{50}$}} &
      \textbf{\makecell[l]{mAP$_{75}$}} &
      \textbf{\makecell[l]{mAP$_{50:95}$}} \\
      \midrule

      \makecell[l]{CIoU terms} &
      \makecell[l]{$\frac{\rho^2}{c^2}$} &
      \makecell[l]{$v \cdot \alpha$} &
      \makecell[l]{40.0} &
      \makecell[l]{29.0} &
      \makecell[l]{26.7} \\

      \makecell[l]{Rescaled aspect term} &
      \makecell[l]{$\frac{\rho^2}{c^2}$} &
      \makecell[l]{$v$} &
      \makecell[l]{40.4} &
      \makecell[l]{28.8} &
      \makecell[l]{26.9} \\

      \makecell[l]{Proposed} &
      \makecell[l]{$\sqrt{\frac{\rho^2}{c^2}}$} &
      \makecell[l]{$v$} &
      \makecell[l]{\textbf{41.0}} &
      \makecell[l]{\textbf{29.2}} &
      \makecell[l]{\textbf{27.3}} \\

      \bottomrule
    \end{tabular}
  }

  \label{tab:ablation_geometry_lse}
\end{table*}

\noindent \textbf{Rescaling the geometric terms.} 
We further rescale the geometric terms to better align with LSE aggregation. In standard CIoU~\cite{ZhengCIoU}, the aspect-ratio term is weighted by $\alpha$, and the center-distance term uses $\frac{\rho^2}{c^2}$. Under LSE, however, the Log-IoU term can dominate due to its larger magnitude. To balance the terms, we remove the $\alpha$ scaling and replace $\frac{\rho^2}{c^2}$ with $\sqrt{\frac{\rho^2}{c^2}}$. Since LSE acts as a soft maximum, better-scaled terms allow each component to contribute more effectively during optimization. Table~\ref{tab:ablation_geometry_lse} shows that this rescaling improves performance.
% We further rescale the geometric terms to make them compatible with LSE aggregation. In standard CIoU~\cite{ZhengCIoU}, the aspect-ratio term is weighted by a dynamic factor $\alpha$, while the center-distance term is used as $\frac{\rho^2}{c^2}$. However, under the LSE framework, the Log-IoU term typically has a larger magnitude and can dominate the aggregation, reducing the effect of the geometric terms. To mitigate this imbalance, we remove the $\alpha$ scaling from the aspect-ratio term and replace the squared center-distance term with $\sqrt{\frac{\rho^2}{c^2}}$. This brings the geometric terms closer in scale to Log-IoU, which is important because LSE behaves as a soft maximum over its components. With better-balanced terms, each component can contribute more meaningfully during optimization. As shown in Table~\ref{tab:ablation_geometry_lse}, this rescaling improves performance and makes the LSE aggregation more effective.

\begin{table}[!ht]
  \centering
  \caption{Ablation study on the weighting strategy in MoEIoU.}
  \resizebox{0.65\textwidth}{!}{%
    \begin{tabular}{lccc}
      \toprule
      \textbf{\makecell[l]{Weights}} &
      \textbf{\makecell[c]{mAP$_{50}$}} &
      \textbf{\makecell[c]{mAP$_{75}$}} &
      \textbf{\makecell[c]{mAP$_{50:95}$}} \\
      \midrule

      \makecell[l]{$w_{\mathrm{iou}}=s$ , $w_{\mathrm{geo}}=1$} &
      \makecell[c]{34.6} & \makecell[c]{24.1} & \makecell[c]{22.6} \\

      \makecell[l]{$w_{\mathrm{iou}}=0.10+0.90s$ , $w_{\mathrm{geo}}=1$} &
      \makecell[c]{34.7} & \makecell[c]{24.2} & \makecell[c]{22.7} \\

      \makecell[l]{$w_{\mathrm{iou}}=s$ , $w_{\mathrm{geo}}=1.00-0.20s$} &
      \makecell[c]{34.6} & \makecell[c]{24.1} & \makecell[c]{22.7} \\

      \makecell[l]{$w_{\mathrm{iou}}=0.10+0.90s$, $w_{\mathrm{geo}}=1.00-0.20s$} &
      \makecell[c]{\textbf{35.0}} & \makecell[c]{\textbf{24.7}} & \makecell[c]{\textbf{23.0}} \\

      \bottomrule
    \end{tabular}
  }

  \label{tab:ablation_weighting_scheme}
\end{table}

\noindent \textbf{Ablation of weighting strategy.} We further ablate the weighting strategy used for the overlap and geometric terms in MoEIoU while keeping the loss terms and LSE aggregation unchanged. Let $s=s_{\mathrm{smoothstep}}(p)$ denote the smoothstep schedule, where $p=t/T$ is the normalized training progress, $t$ is the current epoch, and $T$ is the total number of training epochs. This ablation shows whether the IoU term should start from zero or retain a small early contribution, and whether the geometric terms should remain fixed or mildly decay during training. Table~\ref{tab:ablation_weighting_scheme} shows that the best performance is obtained with
$w_{\mathrm{iou}}=0.10+0.90s$ and $w_{\mathrm{geo}}=1.00-0.20s$. Compared with starting the IoU weight directly from zero, the small initial IoU weight provides early overlap supervision, while the mild decay of the geometric weight gradually shifts emphasis toward precise overlap optimization. 

\begin{table}[!ht]
  \centering
  \caption{Ablation study on aspect-ratio formulation.}
  \resizebox{0.6\textwidth}{!}{%
    \begin{tabular}{lcccc}
      \toprule
      \textbf{\makecell[l]{Formulation}} &
      \textbf{\makecell[l]{Aspect-Ratio Term}} &
      \textbf{\makecell[c]{mAP$_{50}$}} &
      \textbf{\makecell[c]{mAP$_{75}$}} &
      \textbf{\makecell[c]{mAP$_{50:95}$}} \\
      \midrule

      \makecell[l]{EIoU-based} &
      \makecell[l]{$\frac{(w_1-w_2)^2}{w_c^2}+\frac{(h_1-h_2)^2}{h_c^2}$} &
      \makecell[c]{34.8} & \makecell[c]{24.6} & \makecell[c]{22.8} \\

      \makecell[l]{Log-ratio} &
      \makecell[l]{$\left(\log\frac{w_1}{h_1}-\log\frac{w_2}{h_2}\right)^2$} &
      \makecell[c]{34.5} & \makecell[c]{24.4} & \makecell[c]{22.7} \\

      \makecell[l]{Normalized ratio} &
      \makecell[l]{$\left(\frac{\frac{w_1}{h_1}-\frac{w_2}{h_2}}{\frac{w_1}{h_1}+\frac{w_2}{h_2}}\right)^2$} &
      \makecell[c]{34.2} & \makecell[c]{24.5} & \makecell[c]{22.8} \\

      \makecell[l]{CIoU-based} &
      \makecell[l]{$\frac{4}{\pi^2}\left(\arctan\frac{w_2}{h_2}-\arctan\frac{w_1}{h_1}\right)^2$} &
      \makecell[c]{\textbf{35.0}} & \makecell[c]{\textbf{24.7}} & \makecell[c]{\textbf{23.0}} \\

      \bottomrule
    \end{tabular}
  }
  \label{tab:ablation_aspect_term}
\end{table}

\noindent \textbf{Ablation of aspect-ratio formulation} We further study the effect of the aspect-ratio formulation while keeping the overlap and center-distance terms unchanged. 
Here, $(w_1,h_1)$ and $(w_2,h_2)$ denote the predicted and ground-truth box sizes, respectively, and $(w_c,h_c)$ denotes the size of the smallest enclosing box. Table~\ref{tab:ablation_aspect_term} shows that the CIoU-style aspect-ratio term gives the best performance across all metrics.

% Requires:
% \usepackage{pgfplots}
% \pgfplotsset{compat=1.18}

\begin{table*}[!ht]
  \centering
  \caption{Curriculum scheduling ablation in MoEIoU with schedule curves.}
  \label{tab:ablation_curriculum_schedule}

  \makebox[\textwidth][c]{%
  \begin{minipage}[c]{0.4\textwidth}
    \centering
    \resizebox{\linewidth}{!}{%
    \begin{tabular}{lccc}
      \toprule
      \textbf{\makecell[l]{Schedule}} &
      \textbf{\makecell[c]{mAP$_{50}$}} &
      \textbf{\makecell[c]{mAP$_{75}$}} &
      \textbf{\makecell[c]{mAP$_{50:95}$}} \\
      \midrule
      \makecell[l]{Linear}     & \makecell[c]{34.5} & \makecell[c]{24.6} & \makecell[c]{22.7} \\
      \makecell[l]{Cosine~\cite{cosine}}     & \makecell[c]{34.6} & \makecell[c]{24.4} & \makecell[c]{22.7} \\
      \makecell[l]{Sigmoid}    & \makecell[c]{34.6} & \makecell[c]{24.0} & \makecell[c]{22.6} \\
      \makecell[l]{Warmup~\cite{warmup}}     & \makecell[c]{34.6} & \makecell[c]{24.1} & \makecell[c]{22.5} \\
      \makecell[l]{Smoothstep~\cite{smoothstep}} & \makecell[c]{\textbf{35.0}} & \makecell[c]{\textbf{24.7}} & \makecell[c]{\textbf{23.0}} \\
      \bottomrule
    \end{tabular}
    }
  \end{minipage}
  \hspace{-0.02\textwidth}
  \begin{minipage}[c]{0.4\textwidth}
    \centering
    \begin{tikzpicture}
    \begin{axis}[
        width=0.40\linewidth,
        height=0.40\linewidth,
        scale only axis,
        xmin=0, xmax=1,
        ymin=0, ymax=1,
        xlabel={Raw progress $p$},
        ylabel={Scheduled progress $s$},
        grid=major,
        legend style={
    at={(0.98,0.02)},
    anchor=south east,
    font=\fontsize{2}{2.2}\selectfont,
    draw=none,
    fill=white,
    legend columns=1,
    inner sep=0.5pt,
    row sep=-2pt
},
legend image post style={scale=0.35},
        ticklabel style={font=\tiny},
        label style={font=\scriptsize},
        samples=200,
        domain=0:1
    ]

    \definecolor{cbBlue}{RGB}{0,114,178}
    \definecolor{cbGreen}{RGB}{0,158,115}
    \definecolor{cbPurple}{RGB}{204,121,167}
    \definecolor{cbOrange}{RGB}{230,159,0}
    \definecolor{midgray}{RGB}{128,128,128}

    \addplot[very thick, color=cbBlue!90!black] {x};
    \addlegendentry{Linear}

    \addplot[very thick, color=cbPurple!90!black] 
    {(1/(1+exp(-10*(x-0.5))) - 1/(1+exp(-10*(0-0.5)))) /
     ((1/(1+exp(-10*(1-0.5)))) - (1/(1+exp(-10*(0-0.5)))))};
    \addlegendentry{Sigmoid}

    \addplot[very thick, color=cbGreen!90!black] 
    { x <= 0.1 ? 0 : (x-0.1)/(1-0.1) };
    \addlegendentry{Warmup}

    \addplot[very thick, color=midgray!90!black] 
    { x <= 0.1 ? 0 : 0.5 - 0.5*cos(deg(pi*(x-0.1)/(1-0.1))) };
    \addlegendentry{Cosine}

    \addplot[ultra thick, color=cbOrange!90!black] {x^2*(3 - 2*x)};
    \addlegendentry{Smoothstep}

    \end{axis}
    \end{tikzpicture}
  \end{minipage}
  }

  \label{tab:ablation_schedule}
\end{table*}

\noindent \textbf{Ablation of curriculum schedule} 
We also ablate the curriculum schedule while keeping all other loss components fixed. Table~\ref{tab:ablation_curriculum_schedule} shows that the smoothstep schedule achieves the best performance across all metrics. Compared with linear, cosine, sigmoid, and warmup schedules, it provides smoother transitions between overlap and geometric terms, resulting in more stable optimization.
% We also ablate the choice of curriculum schedule by changing the scheduling function, while keeping all other loss components fixed. Table~\ref{tab:ablation_curriculum_schedule} shows that the smoothstep schedule achieves the best results across all metrics. 
% Compared with linear, cosine, sigmoid, and warmup schedules, smoothstep avoids abrupt changes in the relative importance of overlap and geometric terms, leading to more stable optimization.

\section{Conclusion}\label{sec13}
In this work, we introduced MoEIoU, a bounding-box regression loss that adapts to the changing nature of localization errors during training. Instead of using a fixed additive formulation, MoEIoU employs weighted LSE aggregation to dynamically emphasize the most informative regression component for each prediction. A curriculum-based weighting strategy further prioritizes geometric guidance in early training and gradually increases the influence of the IoU term during later refinement stages. Extensive experiments on simulation settings and real detection benchmarks demonstrate the effectiveness of the proposed approach. MoEIoU achieves faster convergence and stronger refinement behavior than existing IoU-based losses, while consistently delivering the best or highly competitive performance across PASCAL VOC, HRIPCB, and the MS COCO subset on multiple YOLO architectures. Ablation studies further confirm the contribution of each component, including LSE aggregation, logarithmic IoU term, curriculum scheduling, and refined geometric formulation.
% In this work, we introduced MoEIoU, a bounding-box regression loss designed to better handle the changing nature of localization errors during training. Instead of combining overlap, center alignment, and aspect-ratio consistency through a fixed additive form, the proposed loss uses a weighted LSE aggregation that adaptively emphasizes the most informative regression component for each prediction. A curriculum-based weighting strategy is further used to place stronger emphasis on geometric guidance in the early stages of training and gradually increase the contribution of the IoU term as training progresses. The proposed MoEIoU was evaluated through both controlled simulation experiments and real detection benchmarks. In the simulation study, MoEIoU showed faster convergence and stronger refinement behavior than existing IoU-based losses. On PASCAL VOC, HRIPCB, and the MS COCO subset, it consistently achieved the best or most competitive performance across different YOLO architectures and evaluation metrics. The ablation studies further confirmed that each design choice, including LSE aggregation, the logarithmic IoU term, curriculum scheduling, and the refined geometric formulation, contributes to the final performance gains.
Overall, the results show that adaptively combining multiple localization signals yields more effective optimization than fixed loss formulations. MoEIoU serves as a simple drop-in replacement for existing IoU-based losses and integrates easily into modern object detectors without architectural changes.
% Overall, the results show that treating bounding-box regression as an adaptive combination of multiple localization signals leads to more effective optimization than relying on fixed loss formulations. MoEIoU provides a simple and practical drop-in replacement for existing IoU-based regression losses and can be integrated into modern object detectors without changing the detector architecture itself. 

%-------------------------------------------------------------------------

\bibliography{egbib}
\clearpage
\onecolumn

\end{document}